# Dyna-Style Planning with Linear Function Approximation and Prioritized Sweeping

**Richard S. Sutton, Csaba Szepesvári, Alborz Geramifard, Michael Bowling**
Reinforcement Learning and Artificial Intelligence Laboratory
Department of Computing Science, University of Alberta, Edmonton, AB Canada T6G 2E8

## Abstract

We consider the problem of efficiently learning optimal control policies and value functions over large state spaces in an online setting in which estimates must be available after each interaction with the world. This paper develops an explicitly model-based approach extending the Dyna architecture to linear function approximation. Dyna-style planning proceeds by generating imaginary experience from the world model and then applying model-free reinforcement learning algorithms to the imagined state transitions. Our main results are to prove that linear Dyna-style planning converges to a unique solution independent of the generating distribution, under natural conditions. In the policy evaluation setting, we prove that the limit point is the least-squares (LSTD) solution. An implication of our results is that prioritized-sweeping can be soundly extended to the linear approximation case, backing up to preceding features rather than to preceding states. We introduce two versions of prioritized sweeping with linear Dyna and briefly illustrate their performance empirically on the Mountain Car and Boyan Chain problems.

## 1 Online learning and planning

Efficient decision making when interacting with an incompletely known world can be thought of as an online learning and planning problem. Each interaction provides additional information that can be used to learn a better model of the world's dynamics, and because this change could result in a different action being best (given the model), the planning process should be repeated to take this into account. However, planning is inherently a complex process; on large problems it not possible to repeat it on every time step without greatly slowing down the response time of the system. Some form of incremental planning is required that, though incomplete on each step, still efficiently computes optimal actions in a timely manner.

The Dyna architecture (Sutton 1990) provides an effective and flexible approach to incremental planning while maintaining responsiveness. There are two ideas underlying the Dyna architecture. One is that planning, acting, and learning are all continual, operating as fast as they can without waiting for each other. In practice, on conventional computers, each time step is shared between planning, acting, and learning, with proportions that can be set arbitrarily according to available resources and required response times.

The second idea underlying the Dyna architecture is that learning and planning are similar in a radical sense. Planning in the Dyna architecture consists of using the model to generate imaginary experience and then processing the transitions of the imaginary experience by model-free reinforcement learning algorithms as if they had actually occurred. This can be shown, under various conditions, to produce exactly the same results as dynamic-programming methods in the limit of infinite imaginary experience.

The original papers on the Dyna architecture and most subsequent extensions (e.g., Singh 1992; Peng & Williams 1993; Moore & Atkeson 1993; Kuvayev & Sutton 1996) assumed a Markov environment with a tabular representation of states. This table-lookup representation limits the applicability of the methods to relatively small problems. Reinforcement learning has been combined with function approximation to make it applicable to vastly larger problems than could be addressed with a tabular approach. The most popular form of function approximation is linear function approximation, in which states or state-action pairs are first mapped to feature vectors, which are then mapped in a linear way, with learned parameters, to value or next-state estimates. Linear methods have been used in many of the successful large-scale applications of reinforcement learning (e.g., Silver, Sutton & Müller 2007; Schaeffer, Hlynka & Jussila 2001). Linear function approximation is also simple, easy to understand, and possesses some of the strongest convergence and performance guarantees among function approximation methods. It is natural then to consider extending Dyna for use with linear function approximation, as we do in this paper.

There has been little previous work addressing planning with linear function approximation in an online setting. Paduraru (2007) treated this case, focusing mainly on sampling stochastic models of a cascading linear form, but also briefly discussing deterministic linear models. Degris, Sigaud and Wuillemin (2006) developed a version of Dyna based on approximations in the form of dynamic Bayes networks and decision trees. Their system, SPITI, included online learning and planning based on an incremental version of structured value iteration (Boutilier, Dearden & Goldszmidt 2000). Singh (1992) developed a version of Dyna for variable resolution but still tabular models. Others have proposed linear least-squares methods for policy evaluation that are efficient in the amount of data used (Bradtke & Barto 1996; Boyan 1999, 2002; Geramifard, Bowling & Sutton 2006). These methods can be interpreted as forming and then planning with a linear model of the world's dynamics, but so far their extensions to the control case have not been well suited to online use (Lagoudakis & Parr 2003; Peters, Vijayakumar & Schaal 2005; Bowling, Geramifard, & Wingate 2008), whereas our linear Dyna methods are naturally adapted to this case. We discuss more specifically the relationship of our work to LSTD methods in a later section. Finally, Atkeson (1993) and others have explored linear, learned models with off-line planning methods suited to low-dimensional continuous systems.

## 2 Notation

We use the standard framework for reinforcement learning with linear function approximation (Sutton & Barto 1998), in which experience consists of the time indexed stream $s_0, a_0, r_1, s_1, a_1, r_2, s_2, \ldots$, where $s_t \in \mathcal{S}$ is a state, $a_t \in \mathcal{A}$ is an action, and $r_t \in \mathbb{R}$ is a reward. The actions are selected by a learning agent, and the states and rewards are selected by a stationary environment. The agent does not have access to the states directly but only through a corresponding feature vector $\phi_t \in \mathbb{R}^n = \phi(s_t)$. The agent selects actions according to a policy, $\pi : \mathbb{R}^n \times \mathcal{A} \to [0,1]$ such that $\sum_{a \in \mathcal{A}} \pi(\phi, a) = 1, \forall \phi$. An important step towards finding a good policy is to estimate the value function for a given policy (policy evaluation). The value function is approximated as a linear function with parameter vector $\theta \in \mathbb{R}^n$:

$$\theta^\top \phi(s) \approx V^\pi(s) = E_\pi \left\{ \sum_{t=1}^{\infty} \gamma^{t-1} r_t \mid s_0 = s \right\},$$

where $\gamma \in [0,1)$. In this paper we consider policies that are greedy or $\epsilon$-greedy with respect to the approximate state-value function.

---

**Algorithm 1**: Linear Dyna for policy evaluation, with random sampling and gradient-descent model learning

Obtain initial $\phi, \theta, F, b$
For each time step:
    Take action $a$ according to the policy. Receive $r, \phi'$
    $\theta \leftarrow \theta + \alpha[r + \gamma \theta^\top \phi' - \theta^\top \phi]\phi$
    $F \leftarrow F + \alpha(\phi' - F\phi)\phi^\top$
    $b \leftarrow b + \alpha(r - b^\top \phi)\phi$
    $temp \leftarrow \phi'$
    Repeat $p$ times (planning):
        Generate a sample $\phi$ from some distribution $\mu$
        $\phi' \leftarrow F\phi$
        $r \leftarrow b^\top \phi$
        $\theta \leftarrow \theta + \alpha[r + \gamma \theta^\top \phi' - \theta^\top \phi]\phi$
    $\phi \leftarrow temp$

---

## 3 Theory for policy evaluation

The natural place to begin a study of Dyna-style planning is with the policy evaluation problem of estimating a state-value function from a linear model of the world. The model consists of a forward transition matrix $F \in \mathbb{R}^n \times \mathbb{R}^n$ (incorporating both environment and policy) and an expected reward vector $b \in \mathbb{R}^n$, constructed such that $F\phi$ and $b^\top \phi$ can be used as estimates of the feature vector and reward that follow $\phi$. A Dyna algorithm for policy evaluation goes through a sequence of planning steps, on each of which a starting feature vector $\phi$ is generated according to a probability distribution $\mu$, and then a next feature vector $\phi' = F\phi$ and next reward $r = b^\top \phi$ are generated from the model. Given this imaginary experience, a conventional model-free update is performed, for example, according to the linear TD(0) algorithm (Sutton 1988):

$$\theta \leftarrow \theta + \alpha(r + \gamma \theta^\top \phi' - \theta^\top \phi)\phi, \quad (1)$$

or according to the residual gradient algorithm (Baird 1995):

$$\theta \leftarrow \theta + \alpha(r + \gamma \theta^\top \phi' - \theta^\top \phi)(\phi - \gamma \phi'), \quad (2)$$

where $\alpha > 0$ is a step-size parameter. A complete algorithm using TD(0), including learning of the model, is given in Algorithm 1.

### 3.1 Convergence and fixed point

There are two salient theoretical questions about the Dyna planning iterations (1) and (2): Under what conditions on $\mu$ and $F$ do they converge? and What do they converge to? Both of these questions turn out to have interesting answers. First, note that the convergence of (1) is in question in part because it is known that linear TD(0) may diverge if the distribution of starting states during training does not match the distribution created by the normal dynamics of the system, that is, if TD(0) is used *off-policy*. This suggests that the sampling distribution used here, $\mu$, might have to be strongly constrained in order for the iteration to be stable. On the other hand, the data here is from the model, and the model is not a general system: it is deterministic[1] and linear. This special case could be much better behaved. In fact, convergence of linear Dyna-style policy evaluation, with either the TD(0) or residual-gradient iterations, is not affected by $\mu$, but only by $F$, as long as $\mu$ exercises all directions in the full $n$-dimensional vector space. Moreover, not only is the fact of convergence unaffected by $\mu$, but so is the value converged to. In fact, we show below that convergence is to a deterministic fixed point, a value of $\theta$ such that the iterations (1) and (2) leave it unchanged not just in expected value, but for every individual $\phi$ that could be generated by $\mu$. The only way this could be true is if the TD error (the first expression in parentheses in each iteration) were exactly zero, that is, if

$$\begin{aligned} 0 &= r + \gamma \theta^\top \phi' - \theta^\top \phi \\ &= b^\top \phi + \gamma \theta^\top F \phi - \theta^\top \phi \\ &= (b + \gamma F^\top \theta - \theta)^\top \phi. \end{aligned}$$

And the only way that this can be true for all $\phi$ is for the expression in parenthesis above to be zero:

$$\begin{aligned} 0 &= b + \gamma F^\top \theta - \theta \\ &= b + (\gamma F^\top - I)\theta, \end{aligned}$$

which immediately implies that

$$\theta = (I - \gamma F^\top)^{-1} b, \qquad (3)$$

assuming that the inverse exists. Note that this expression for the fixed point does not depend on $\mu$, as promised.

If $I - \gamma F^\top$ is nonsingular, then there might be no fixed point. This could happen for example if $F$ were an expansion, or more generally if the limit $(\gamma F)^\infty$ were not zero. These cases correspond to world models that say the feature vectors diverge to infinity over time. Failure to converge in these cases should not be considered a problem for the Dyna iterations as planning algorithms; these are cases in which the planning *problem* is ill posed. If the feature vectors diverge, then so too may the rewards, in which case the true values given the model are infinite. No real finite Markov decision process could behave in this way.

It remains to show the conditions on $F$ under which the iterations converge to the fixed point if one exists. We prove next that under the TD(0) iteration (1), convergence is guaranteed if the numerical radius of $F$ is less than one,[2] and then that under the residual-gradient iteration (2), convergence is guaranteed for any $F$ as long as the fixed point exists. That $F$'s numerical radius be less than 1 is a stronger condition than nonsingularity of $I - \gamma F^\top$, but it is similar in that both conditions pertain to the matrix trending toward expansion when multiplied by itself.

---

[1]The model is deterministic because it generates the expectation of the next feature vector; the system itself may be stochastic.

[2]The numerical radius of a real-valued square matrix $A$ is defined by $r(A) = \max_{\|x\|_2 = 1} x^T A x$.

**Theorem 3.1** (Convergence of linear TD(0) Dyna for policy evaluation). *Consider the TD(0) iteration with a nonnegative step-size sequence* $(\alpha_k)$:

$$\theta_{k+1} = \theta_k + \alpha_k (b^\top \phi_k + \gamma \theta_k^\top F \phi_k - \theta_k^\top \phi_k)\phi_k, \qquad (4)$$

*where $\theta_0 \in \mathbb{R}^n$ is arbitrary. Assume that (i) the step-size sequence satisfies $\sum_{k=0}^\infty \alpha_k = \infty$, $\sum_{k=0}^\infty \alpha_k^2 < \infty$, (ii) $r(F) \leq 1$, (iii) $(\phi_k)$ are uniformly bounded i.i.d. random variables, and that (iv) $C = \mathbb{E}\left[\phi_k \phi_k^\top\right]$ is non-singular. Then the parameter vector $\theta_k$ converges with probability one to $(I - \gamma F^\top)^{-1} b$.*

*Proof.* The idea of the proof is to view the algorithm as a stochastic gradient descent method. In particular, we apply Proposition 4.1 of (Bertsekas & Tsitsiklis 1996).

Before verifying the conditions of this result, let us rewrite (4) in terms of the matrix $G = I - \gamma F$:

$$\begin{aligned} \theta_{k+1} &= \theta_k + \alpha_k (b^\top \phi_k + \theta_k^\top (\gamma F - I)\phi_k)\phi_k \\ &= \theta_k + \alpha_k (b^\top \phi_k - \theta_k^\top G \phi_k)\phi_k \\ &= \theta_k + \alpha_k s_k. \end{aligned}$$

Here $s_k$ is defined by the last equation.

The cited proposition requires the definition of a potential function $J(\theta)$ and will allow us to conclude that $\lim_{k \to \infty} \nabla J(\theta_k) = 0$ with probability one. Let us choose $J(\theta) = 1/2 \, \mathbb{E}\left[(b^\top \phi_k + \gamma \theta^\top F \phi_k - \theta^\top \phi_k)^2\right]$. Note that by our i.i.d. assumptions on the features, $J(\theta)$ is well-defined. We need to check four conditions (because the step-size conditions are automatically satisfied): (i) The nonnegativity of the potential function; (ii) The Lipschitz continuity of $\nabla J(\theta)$; (iii) The pseudo-gradient property of the expected update direction; and (iv) The boundedness of the expected magnitude of the update, more precisely that $\mathbb{E}\left[\|s_k\|_2^2 | \theta_k\right] \leq O(\|\nabla J(\theta_k)\|_2^2)$. Nonnegativity is satisfied by definition and the boundedness condition (iv) is satisfied thanks to the boundedness of the features.

Let us show now that the pseudo-gradient property (iii) is satisfied. This condition requires the demonstration of a positive constant $c$ such that

$$c\|\nabla J(\theta_k)\|_2^2 \leq -\nabla J(\theta_k)^\top \mathbb{E}\left[s_k | \theta_k\right]. \qquad (5)$$

Define $\bar{s}_k = \mathbb{E}\left[s_k | \theta_k\right] = Cb - CG^\top \theta_k$. A simple calculation gives $\nabla J(\theta_k) = -G\bar{s}_k$. Hence $\|\nabla J(\theta_k)\|_2^2 = \bar{s}_k^\top G^\top G \bar{s}_k$ and $-(\nabla J(\theta_k))^\top \bar{s}_k = \bar{s}_k^\top G \bar{s}_k$. Therefore (5) is equivalent to $c\bar{s}_k^\top G^\top G \bar{s}_k \leq \bar{s}_k^\top G \bar{s}_k$. In order to make this true with a sufficiently small $c$, it suffices to show that

$s^\top G s > 0$ holds for any non-zero vector $s$. An elementary reasoning shows that this is equivalent to $1/2(G+G^\top)$ being positive definite, which in turn is equivalent to $r(F) \leq 1$, showing that (iii) is satisfied.

Hence, we have verified all the assumptions of the cited proposition and can therefore we conclude that $\lim_{k\to\infty} \nabla J(\theta_k) = 0$ with probability one. Plugging in the expression of $\nabla J(\theta_k)$, we get $\lim_{t\to\infty}(Cb - CG^\top \theta_k) = 0$. Because $C$ and $G$ are invertible (this latter follows from $r(F) \leq 1$), it follows that the limit of $\theta_k$ exists and $\lim_{k\to\infty} \theta_k = (G^\top)^{-1} b = (I - \gamma F^\top)^{-1} b$. □

Several extensions of this result are possible. First, the requirement of i.i.d. sampling can be considerably relaxed. With an essentially unchanged proof, it is possible to show that the theorem remains true if the feature vectors are generated by a Markov process given that they satisfy appropriate ergodicity conditions. Moreover, building on a result by Delyon (1996), one can show that the result continues to hold even if the sequence of features is generated in an algorithmic manner, again provided that some ergodicity conditions are met. The major assumption then is that $C = \lim_{K\to\infty} 1/K \sum_{k=1}^{K} \phi_k \phi_k^\top$ exists and is non-singular. Further, because there is no "noise" to reject, there is no need to decay the step-sizes towards zero (the condition $\sum_{k=0}^{\infty} \alpha_k^2 < +\infty$ in the proofs is used to "filter out noise"). In particular, we conjecture that sufficiently small constant step-sizes would work as well (for a result of this type see Proposition 3.4 by Bertsekas & Tsitsiklis 1996).

On the other hand the requirement on the numerical radius of $F$ seems to be necessary for the convergence of the TD(0) iteration. By studying the ODE associated with (4), we see that it is stable if and only if $CG$ is a positive stable matrix (i.e., iff all its eigenvalues have positive real part). From this it seems necessary to require that $G$ is positive stable. However, to ensure that $CG$ is positive stable the strictly stronger condition that $G + G^\top$ is positive definite must be satisfied. This latter condition is equivalent to $r(F) \leq 1$.

We turn now to consider the convergence of Dyna planning using the residual-gradient Dyna iteration (2). This update rule can be derived by taking the gradient of $J(\theta, \phi_k) = (b^\top \phi_k + \gamma \theta^\top \phi_k' - \theta^\top \phi_k)^2$ w.r.t. $\theta$. Thus, as an immediate consequence of Proposition 4.1 of (Bertsekas & Tsitsiklis 1996) we get the following result:

**Theorem 3.2** (Convergence of residual-gradient Dyna for policy evaluation). *Assume that $\theta_k$ is updated according to*

$$\theta_{k+1} = \theta_k + \alpha_k (b^\top \phi_k + \gamma \theta_k^\top F \phi_k - \theta_k^\top \phi_k)(\phi_k - \gamma F \phi_k),$$

*where $\theta_0 \in \mathbb{R}^n$ is arbitrary. Assume that the non-negative step-size sequence $(\alpha_k)$ satisfies the summability condition (i) of Theorem 3.1 and that $(\phi_k)$ are uniformly bounded i.i.d. random variables. Then the parameter vector $\theta_k$ converges with probability one to $(I - \gamma F^\top)^{-1} b$, assuming that $(I - \gamma F^\top)$ is non-singular.*

*Proof.* As all the conditions of Proposition 4.1 of (Bertsekas & Tsitsiklis 1996) are trivially satisfied with the choice $J(\theta) = \mathbb{E}[J(\theta, \phi_k)]$, we can conclude that $\theta_k$ converges w.p.1 to the minimizer of $J(\theta)$. In the previous theorem we have seen that the minimizer of $J(\theta)$ is indeed $\theta = (I - \gamma F^\top)^{-1} b$, finishing the proof. □

### 3.2 Convergence to the LSTD solution

So far we have discussed the convergence of planning given a model, but we have said nothing about the relationship of the model to data, or about the quality of the resultant solution. Suppose the model were the best linear fit to a finite dataset of observed feature-vector-to-feature-vector transitions with accompanying rewards. In this case we can show that the fixed point of the Dyna updates is the least squares temporal-difference solution. This is the solution for which the mean TD(0) update is zero and is also the solution found by the LSTD(0) algorithm (Barto & Bradtke 1996).

**Theorem 3.3.** *Given a training dataset of feature, reward, next-state feature triples $D = [\phi_1, r_1, \phi_1', \ldots, \phi_n, r_n, \phi_n']$, let $F, b$ be the least-squares model built on $D$. Assume that $C = \sum_{k=1}^{n} \phi_k \phi_k^\top$ has full rank. Then the solution (3) is the same as the LSTD solution on this training set.*

*Proof.* It suffices to show that the respective solution sets of the equations

$$0 = \sum_{k=1}^{n} \phi_k (r_k + \gamma (\phi_k')^\top \theta - \phi_k^\top \theta), \quad (6)$$

$$0 = b + (\gamma F^\top - I)\theta \quad (7)$$

are the same. This is because the LSTD parameter vectors are obtained by solving the first equation and the TD(0) Dyna solutions are derived from the second equation.

Let $D = \sum_{k=1}^{n} \phi_k (\phi_k')^\top$, and $\bar{r} = \sum_{k=1}^{n} \phi_k r_k$. A standard calculation shows that

$$F^\top = C^{-1} D \quad \text{and} \quad b = C^{-1} \bar{r}.$$

Plugging in $C$, $D$ into (6) and factoring out $\theta$ shows that any solution of (6) also satisfies

$$0 = \bar{r} + (\gamma D - C)\theta. \quad (8)$$

If we multiply both sides of (8) by $C^{-1}$ from the left we get (7). Hence any solution of (6) is also a solution of (7). Because all the steps of the above derivation are reversible, we get that the reverse statement holds as well. □

**Algorithm 2** : Linear Dyna with PWMA prioritized sweeping (policy evaluation)

Obtain initial $\phi, \theta, F, b$
For each time step:
    Take action $a$ according to the policy. Receive $r, \phi'$
    $\delta \leftarrow r + \gamma \theta^\top \phi' - \theta^\top \phi$
    $\theta \leftarrow \theta + \alpha \delta \phi$
    $F \leftarrow F + \alpha(\phi' - F\phi)\phi^\top$
    $b \leftarrow b + \alpha(r - b^\top \phi)\phi$
    For all $i$ such that $\phi(i) \neq 0$:
        For all $j$ such that $F^{ij} \neq 0$:
            Put $j$ on the PQueue with priority $|F^{ij}\delta\phi(i)|$
    Repeat $p$ times while PQueue is not empty:
        $i \leftarrow$ pop the PQueue
        $\delta \leftarrow b(i) + \gamma \theta^\top F e_i - \theta(i)$
        $\theta(i) \leftarrow \theta(i) + \alpha \delta$
        For all $j$ such that $F^{ij} \neq 0$:
            Put $j$ on the queue with priority $|F^{ij}\delta|$
    $\phi \leftarrow \phi'$

**Algorithm 3** : Linear Dyna with MG prioritized sweeping (policy evaluation)

Obtain initial $\phi, \theta, F, b$
For each time step:
    Take action $a$ according to the policy. Receive $r, \phi'$
    $\delta \leftarrow r + \gamma \theta^\top \phi' - \theta^\top \phi$
    $\theta \leftarrow \theta + \alpha \delta \phi$
    $F \leftarrow F + \alpha(\phi' - F\phi)\phi^\top$
    $b \leftarrow b + \alpha(r - b^\top \phi)\phi$
    For all $i$ such that $\phi(i) \neq 0$:
        Put $i$ on the PQueue with priority $|\delta\phi(i)|$
    Repeat $p$ times while PQueue is not empty:
        $i \leftarrow$ pop the PQueue
        For all $j$ such that $F^{ij} \neq 0$:
            $\delta \leftarrow b(j) + \gamma \theta^\top F e_j - \theta(j)$
            $\theta(j) \leftarrow \theta(j) + \alpha \delta$
            Put $j$ on the PQueue with priority $|\delta|$
    $\phi \leftarrow \phi'$

## 4 Linear prioritized sweeping

We have shown that the convergence and fixed point of policy evaluation by linear Dyna are not affected by the way the starting feature vectors are chosen. This opens the possibility of selecting them cleverly so as to speed the convergence of the planning process. One natural idea—the idea behind prioritized sweeping—is to work backwards from states that have changed in value to the states that lead into them. The lead-in states are given priority for being updated because an update there is likely to change the state's value (because they lead to a state that has changed in value). If a lead-in state is updated and its value is changed, then *its* lead-in states are in turn given priority for updating, and so on. In the table-lookup context in which this idea was developed (Moore & Atkeson 1993; Peng 1993; see also Wingate & Seppi 2005), there could be many states preceding each changed state, but only one could be updated at a time. The states waiting to be updated were kept in a queue, prioritized by the size of their likely effect on the value function. As high-priority states were popped off the queue and updated, it would sometimes give rise to highly efficient sweeps of updates across the state space; this is what gave rise to the name "prioritized sweeping".

With function approximation it is not possible to identify and work backwards from individual states, but alternatively one could work backwards feature by feature. If there has just been a large change in $\theta(i)$, the component of the parameter vector corresponding to the $i$th feature, then one can look backwards through the model to find the features $j$ whose components $\theta(j)$ are likely to have changed as a result. These are the features $j$ for which the elements $F^{ij}$ of $F$ are large. One can then preferentially construct starting feature vectors $\phi$ that have non-zero entries at these $j$ components. In our algorithms we choose the starting vectors to be the unit basis vectors $e_j$, all of whose components are zero except the $j$th, which is 1. (Our theoretical results assure us that this cannot affect the result of convergence.) Using unit basis vectors is very efficient computationally, as the vector matrix multiplication $F\phi$ is reduced to pulling out a single column of $F$.

There are two tabular prioritized sweeping algorithms in the literature. The first, due simultaneously to Peng and Williams (1993) and to Moore and Atkeson (1993), which we call *PWMA prioritized sweeping*, adds the predecessors of every state encountered in real experience to the priority queue whether or not the value of the encountered state was significantly changed. The second form of prioritized sweeping, due to McMahan and Gordon (2005), and which we call *MG prioritized sweeping*, puts each encountered state on the queue, but not its predecessors. For McMahan and Gordon this resulted in a more efficient planner. A complete specification of our feature-by-feature versions of these two forms of prioritized sweeping are given above, with TD(0) updates and gradient-descent model learning, as Algorithms 2 and 3. These algorithms differ slightly from previous prioritized sweeping algorithms in that they update the value function from the real experiences and not just from model-generated experience. With function approximation, real experience is always more informative than model-generated experience, which will be distorted by the function approximator. We found this to be a significant effect in our empirical experiments (Section 6).

**Algorithm 4**: Linear Dyna with MG prioritized sweeping and TD(0) updates (control)

---
Obtain initial $\phi, \theta, F, b$
For each time step:
    $a \leftarrow \arg\max_a \left[ b_a^\top \phi + \gamma \theta^\top F_a \phi \right]$      (or $\epsilon$-greedy)
    Take action $a$, receive $r, \phi'$
    $\delta \leftarrow r + \gamma \theta^\top \phi' - \theta^\top \phi$
    $\theta \leftarrow \theta + \alpha \delta \phi$
    $F_a \leftarrow F_a + \alpha(\phi' - F_a\phi)\phi^\top$
    $b_a \leftarrow b_a + \alpha(r - b_a^\top \phi)\phi$
    For all $i$ such that $\phi(i) \neq 0$:
        Put $i$ on the PQueue with priority $|\delta \phi(i)|$
    Repeat $p$ times while PQueue is not empty:
        $i \leftarrow$ pop the PQueue
        For all $j$ s.t. there exists an $a$ s.t. $F_a^{ij} \neq 0$:
            $\delta \leftarrow \max_a \left[ b_a(j) + \gamma \theta^\top F_a e_j \right] - \theta(j)$
            $\theta(j) \leftarrow \theta(j) + \alpha \delta$
            Put $j$ on the PQueue with priority $|\delta|$
    $\phi \leftarrow \phi'$

---

## 5 Theory for Control

We now turn to the full case of control, in which separate models $F_a, b_a$ are learned and are then available for each action $a$. These are constructed such that $F_a \phi$ and $b_a^\top \phi$ can be used as estimates of the feature vector and reward that follow $\phi$ if action $a$ is taken. A linear Dyna algorithm for the control case goes through a sequence of planning steps on each of which a starting feature vector $\phi$ and an action $a$ are chosen, and then a next feature vector $\phi' = F_a \phi$ and next reward $r = b_a \phi$ are generated from the model. Given this imaginary experience, a conventional model-free update is performed. The simplest case is to again apply (1). A complete algorithm including prioritized sweeping is given in Algorithm 4.

The theory for the control case is less clear than for policy evaluation. The main issue is the stability of the "mixture" of the forward model matrices. The corollary below is stated for an i.i.d. sequence of features, but by the remark after Theorem 3.1 it can be readily extended to the case where the policy to be evaluated is used to generate the trajectories.

**Corollary 5.1** (Convergence of linear TD(0) Dyna with action models)**.** *Consider the Dyna recursion (4) with the modification that in each step, instead of $F\phi_k$, we use $F_{\pi(\phi_k)}\phi_k$, where $\pi$ is a policy mapping feature vectors to actions and $\{F_a\}$ is a collection of forward-model matrices. Similarly, $b^\top \phi_k$ is replaced by $b_{\pi(\phi_k)}^\top \phi_k$. As before, assume that $\phi_k$ is an unspecified i.i.d. process. Let $(F, b)$ be the least squares model of $\pi$: $F = \arg\min_G \mathbb{E}\left[ \|G\phi_k - F_{\pi(\phi_k)}\phi_k\|_2^2 \right]$ and $b = \arg\min_u \mathbb{E}\left[ (u^\top \phi_k - b_{\pi(\phi_k)}^\top \phi_k)^2 \right]$ If the numerical radius of $F$ is bounded by one, then the conclusions of Theorem 3.1 hold: the parameter vector $\theta_k$ converges with probability one to $(I - \gamma F^\top)^{-1} b$.*

*Proof.* The proof is immediate from the normal equation for $F$, which states that $\mathbb{E}\left[ F \phi_k \phi_k^\top \right] = \mathbb{E}\left[ F_{\pi(\phi_k)} \phi_k \phi_k^\top \right]$, and once we observe that, in the proof of Theorem 3.1, $F$ appears only in expressions of the form $\mathbb{E}\left[ F \phi_k \phi_k^\top \right]$. $\square$

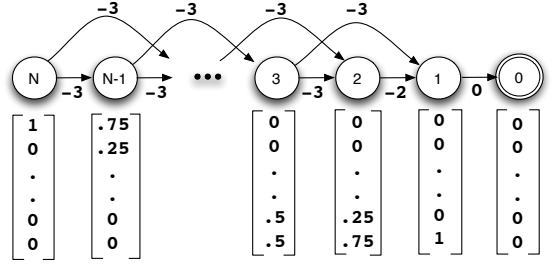

Figure 1: The general Boyan Chain problem.

As in the case of policy evaluation, there is a corresponding corollary for the residual gradient iteration, with an immediate proof. These corollaries say that, for any policy with a corresponding model that is stable, the Dyna recursion can be used to compute its value function. Thus we can perform a form of policy iteration—continually computing an approximation to the value function for the greedy policy.

## 6 Empirical results

In this section we illustrate the empirical behavior of the four Dyna algorithms and make comparisons to model-free methods using variations of two standard test problems: Boyan Chain and Mountain Car. Our Boyan Chain environment is an extension of that by Boyan (1999, 2002) from 13 to 98 states, and from 4 to 25 features (Geramifard, Bowling & Sutton 2006). Figure 1 depicts this environment in the general form. Each episode starts at state $N = 98$ and terminates in state 0. For all states $s > 2$, there is an equal probability of transitioning to states $s - 1$ or $s - 2$ with a reward of $-3$. From states 2 and 1, there are deterministic transitions to states 1 and 0 with respective rewards of $-2$ and $0$. Our Mountain Car environment is exactly as described by Sutton (1996; Sutton & Barto 1998), re-implemented in Matlab. An underpowered car must be driven to the top of a hill by rocking back and forth in a valley. The state variables are a pair (position,velocity) initialized to $(-0.5, 0.0)$ at the beginning of each episode. The reward is $-1$ per time step. There are three discrete actions (accelerate, reverse, and coast). We used a value function representation based on tile-coding feature vectors exactly as in Sutton's (1996) experiments, with 10 tilings over the combined (position, velocity) pair, and with the tiles hashed down to 10,000 features. In the policy evaluation experiments with this domain, the policy was to accelerate in the direction of the current velocity, and we added noise to the domain that switched the selected action to a random action with 10% probability. Complete code for our test problems as standard RL-Glue environments is available from the RL-Library hosted at the University of Alberta.

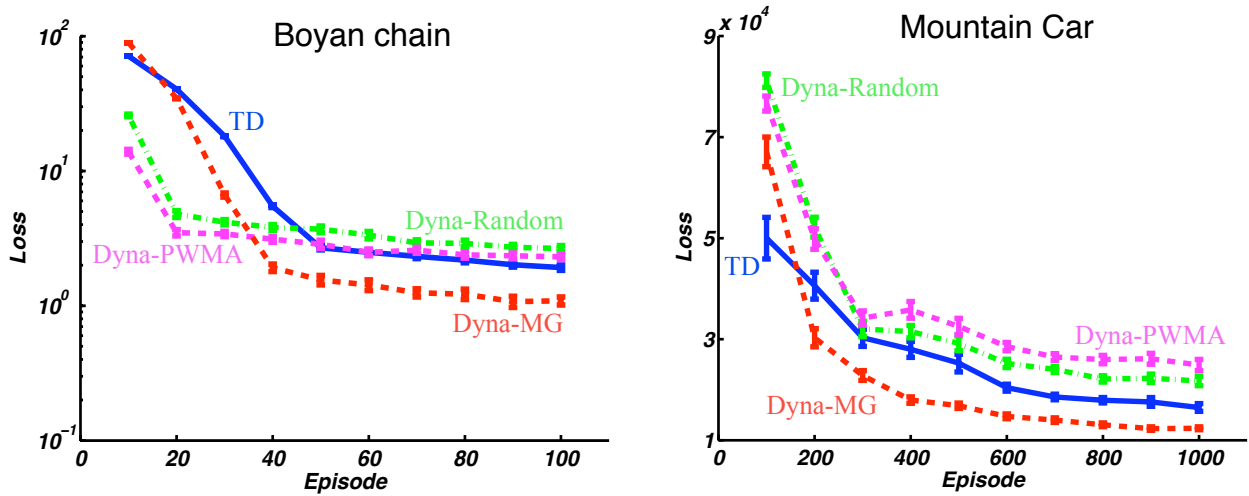

Figure 2: Performance of policy evaluation methods on the Boyan Chain and Mountain Car environments

In all experiments, the step size parameter $\alpha$ took the form $\alpha_t = \alpha_0 \frac{N_0+1}{N_0+t^{1.1}}$, in which $t$ is the episode number and the pair $(N_0, \alpha_0)$ was selected based on empirically finding the best combination out of $\alpha_0 \in \{.01, .1, 1\}$ and $N_0 \in \{100, 1000, 10^6\}$ separately for each algorithm and domain. All methods observed the same trajectories in policy evaluation. All graphs are averages of 30 runs; error bars indicate standard errors in the means. Other parameter settings were $\epsilon = 0.1$, $\gamma = 1$, and $\lambda = 0$.

We performed policy evaluation experiments with four algorithms: Dyna-Random, Dyna-PWMA, Dyna-MG (as in Algorithms 1–3), and model-free TD(0). In the case of the Dyna-Random algorithm, the starting feature vectors in planning were chosen to be unit basis vectors with the 1 in a random location. Figure 2 shows the policy evaluation performance of the four methods in the Boyan Chain and Mountain Car environments. For the Boyan Chain domain, the loss was the root-mean-squared error of the learned value function compared to the exact analytical value, averaged over all states. In the Mountain Car domain, the states are visited very non-uniformly, and a more sophisticated measure is needed. Note that all of the methods drive $\theta$ toward an asymptotic value in which the expected TD(0) update is zero; we can use the distance from this as a loss measure. Specifically, we evaluated each learned value function by freezing it and then running a fixed set of 200,000 episodes with it while running the TD(0) algorithm (but not allowing $\theta$ to actually change). The norm of the sum of the (attempted) update vectors was then computed and used as the loss. In practice, this measure can be computed very efficiently as $||A^*\theta - b^*||$ (in the notation of LSTD(0), see Bradtke & Barto 1996).

In the Boyan Chain environment, the Dyna algorithms generally learned more rapidly than model-free TD(0). Dyna-MG was initially slower than the other algorithms, then caught up and surpassed them. The relatively poor early performance of Dyna-MG was actually due to its being a better planning method. After few episodes the model tends to be of very high variance, and so therefore is the best value-function estimate given it. We tested this hypothesis by running the Dyna methods starting with a fixed, well-learned model; in this case Dyna-MG was the best of all the methods from the beginning. All of these data are for one step of planning for each real step of interaction with the world ($p = 1$). In preliminary experiments with larger values of $p$, up to $p = 10$, we found further improvements in learning rate of the Dyna algorithms over TD(0), and again Dyna-MG was best.

The results for Mountain Car are less clear. Dyna-MG quickly does significantly better than TD(0), but the other Dyna algorithms lag initially and never surpass TD(0). Note that, for any value of $p$, Dyna-MG does many more $\theta$ updates than the other two Dyna algorithms (because these updates are in an inner loop, cf. Algorithms 2 and 3). Even so, because of its other efficiencies Dyna-MG tended to run faster overall in our implementation. Obviously, there is a lot more interesting empirical work that could be done here.

We performed one Mountain Car experiment with Dyna-MG as a *control* algorithm (Algorithm 4), comparing it with model-free Sarsa (i.e., Algorithm 4 with $p = 0$). The results are shown in Figure 3. As before, Dyna-MG showed a distinct advantage over the model-free method in terms of learning rate. There was no clear advantage for either method in the second half of the experiment. We note that, asymptotically, model-free methods are never worse than model-based methods, and are often better because the model does not converge exactly to the true system because of structural modeling assumptions. (The case we treat here—linear models and value functions with one-step TD methods—is a rare case in which asymptotic performance of model-based and model-free methods should be identical.) The benefit of models, and of planning generally, is in rapid adaptation to new problems and situations.

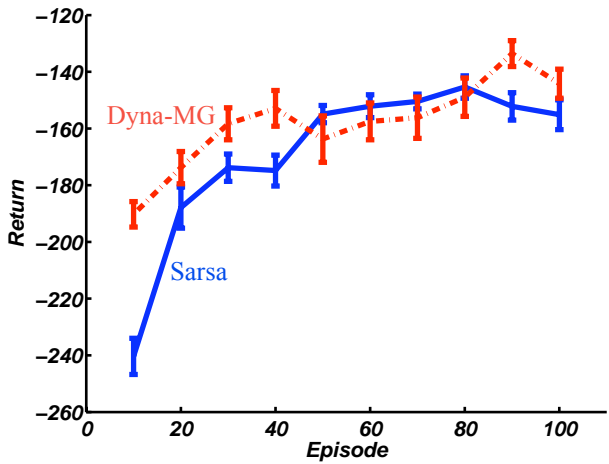

Figure 3: Control performance on Mountain Car

These empirical results are not extensive and in some cases are preliminary, but they nevertheless illustrate some of the potential of linear Dyna methods. The results on the Boyan Chain domain show that Dyna-style planning can result in a significant improvement in learning speed over model-free methods. In addition, we can see trends that have been observed in the tabular case re-occurring here with linear function approximation. In particular, prioritized sweeping can result in more efficient learning than simply updating features at random, and the MG version of prioritized sweeping seems to be better than the PWMA version.

Finally, we would like to note that we have done extensive experimental work (not reported here) attempting to adapt least squares methods such as LSTD to online control domains, in particular to the Mountain Car problem. A major difficulty with these methods is that they place equal weight on all past data whereas, in a control setting, the policy changes and older data becomes less relevant and may even be misleading. Although we have tried a variety of forgetting strategies, it is not easy to obtain online control performance with these methods that is superior to model-free methods. One reason we consider the Dyna approach to be promising is that no special changes are required for this case; it seems to adapt much more naturally and effectively to the online control setting.

## 7 Conclusion

In this paper we have taken important steps toward establishing the theoretical and algorithmic foundations of Dyna-style planning with linear function approximation. We have established that Dyna-style planning with familiar reinforcement learning update rules converges under weak conditions corresponding roughly, in some cases, to the existence of a finite solution to the planning problem, and that convergence is to a unique least-squares solution independent of the distribution used to generate hypothetical experience. These results make possible our second main contribution: the introduction of algorithms that extend prioritized sweeping to linear function approximation, with correctness guarantees. Our empirical results illustrate the use of these algorithms and their potential for accelerating reinforcement learning. Overall, our results support the conclusion that Dyna-style planning may be a practical and competitive approach to achieving rapid, online control in stochastic sequential decision problems with large state spaces.

**Acknowledgements**

The authors gratefully acknowledge the substantial contributions of Cosmin Paduraru and Mark Ring to the early stages of this work. This research was supported by iCORE, NSERC and Alberta Ingenuity.